 \documentclass[pmlr,twocolumn,10pt]{jmlr} 
\PassOptionsToPackage{linesnumbered, ruled, vlined, noresetcount,algosection}{algorithm2e}

\setcitestyle{square, comma, numbers,sort&compress, super}

\usepackage{booktabs}
\usepackage[load-configurations=version-1]{siunitx} 
\usepackage{multirow}
\usepackage{xcolor}
\usepackage{graphics}
\usepackage{gensymb}
\usepackage{dblfloatfix}
\usepackage{tikz}
\usepackage{amsmath}
\usepackage{mathtools}
\usepackage{caption}
\usepackage{multirow}
\usepackage{url}
\usepackage{hyperref}
\usepackage{etoolbox}
\usepackage{chngcntr}
\usepackage[ruled, vlined]{algorithm2e}

\makeatletter
    \let\c@algocf\c@subsection
\makeatother

\renewcommand{\thealgocf}
 {{\thesection}.\arabic{algocf}}

\theorembodyfont{\upshape}
\theoremheaderfont{\scshape}
\theorempostheader{:}
\theoremsep{\newline}

\jmlrvolume{}
\jmlryear{2021}
\jmlrpublished{ML4H Extended Abstract Arxiv Index}
\jmlrworkshop{NeurIPS 2021 Workshop} 
\title[Shared Model for Predicting Motorized Prosthetic Joint Motion]{Learning a Shared Model for Motorized Prosthetic Joints to Predict Ankle-Joint Motion}

   \author{%
\Name{Sharmita Dey} \Email{Sharmita.dey@med.uni-goettingen.de}\\
\addr University of Goettingen, Germany
\AND
\Name{Sabri Boughorbel} \Email{sboughorbel@hbku.edu.qa}\\
\addr Qatar Computing Research Institute, HBKU, Qatar
\AND
\Name{Arndt F. Schilling} \Email{Arndt.Schilling@med.uni-goettingen.de}\\
\addr University Medical Center Goettingen, Germany
}
\begin{document}

\maketitle

\begin{abstract}
Control strategies for active prostheses/orthoses
use sensor inputs to recognize the user’s locomotive intention and generate corresponding control commands for producing the desired locomotion. In this paper, we propose a learning-based shared model  for predicting ankle-joint motion for different locomotion modes like level-ground walking, stair ascent, stair descent, slope ascent, and slope descent without the need to classify between them. Features extracted from hip and knee joint angular motion are used to continuously predict the ankle angles and moments using a Feed-Forward Neural Network-based shared model. We show that the shared model is adequate for predicting the ankle angles and moments for different locomotion modes without explicitly classifying between the modes. The proposed strategy shows the potential for devising a high-level controller for an intelligent prosthetic ankle that can adapt to different locomotion modes.

\end{abstract}

\section{Introduction}
Lower limb amputations and stroke affects the ability of normal locomotion. In order to restore the lost or disabled locomotive functionality, amputees and stroke survivors have to use assistive devices like prostheses and orthoses. Most of the commercially available  prosthetic/orthotic devices offers only passive assistance 
that do not produce adequate energy for a normal gait pattern \cite{windrich2016active}. The additional energy required for locomotion has to be compensated by the intact/residual body parts which can lead to increased metabolic cost and asymmetric gait patterns, and overload on the residual joints \cite{au2008powered}. On the other hand, active or powered prostheses can produce mechanical power using motors and thus have the potential to allow more natural locomotion \cite{au2008powered}.  
With the advancements in computational technology and processing power, the notion of intelligent powered prosthetic/orthotic devices has become more realistic \cite{Inpro}. Such intelligent prosthetic/orthotic devices are expected to understand the user's intention using sensor input and assist in performing the required task to replace the lost functionality. The control algorithm for an intelligent prosthesis should ideally be able to automatically adapt to different locomotion modes or terrains and different locomotion speeds. 
The control of active prostheses is commonly implemented using a multilevel control approach \cite{varol2010multiclass, Tucker2015}.

High level control strategies can be broadly divided into two categories: multi-stage models and direct control models. In the multi-stage models, sensory inputs are used to classify between different gait modes and switch between the mid-level controllers designated for each mode. Different sensory inputs like the body joint angular positions  \cite{dey2020continuous}, surface electromyography (EMG) signals \cite{huang2009strategy} or a fusion of different sensor data \cite{huang2011continuous, Tkach2013Neuromechanical} have been used. Different studies have proposed the use of pattern recognition algorithms like linear discriminant analysis classifiers \cite{ young2013classifying}, Gaussian Mixture Models \cite{varol2010multiclass}, dynamic Bayesian networks \cite{young2014intent} and artificial neural networks \cite{feng2019strain} for locomotion mode recognition.

On the other hand, the direct control approach uses the user's intention deduced from the sensory inputs from user's residual body parts to directly alter the state of the prosthetic/orthotic device. Electromyography (EMG) signals from the user's residual muscles \cite{coker2021emg} or mechanical inputs, like, IMU data \cite{mundt2020estimation} are used to directly modulate the angle or moment of the prosthetic/orthotic actuator to produce the desired locomotion.  
While the direct control approach can potentially prevent the need for classification between different locomotion modes, most of the existing studies support  single locomotion mode (e.g., level walking \cite{coker2021emg, dey2020feasibility, lim2020prediction} or stair ascent \cite{hoover2013stair}), or non-weight bearing situations like joint movements while sitting \cite{hargrove2013non}. \\

The aim of this work is to develop a learning-based shared model for a powered prosthetic ankle which can continuously predict the required angles and moments across different locomotion tasks/modes without the need for explicitly classifying between the them.

\section{Methodology}

\subsection{Motion Capture Experiments} \label{biomech-analysis}
Three-dimensional motion data were recorded using a motion capture system (VICON Motion Systems, Ltd., UK), from an able-bodied male subject while he performed different locomotion trials: level ground walking at self-selected normal (comfortable) speeds, stair ascent and descent, and slope ascent and descent at an inclination of 10 degrees. The motion data were captured at a rate of 200 Hz. Ten trials were recorded for level ground walking at a self-selected normal speed. Later, due to time limits, eight trials were recorded for stair ascent, stair descent, and slope ascent. Although eight trials were recorded for slope descent, one trial had to be discarded for data corruption. Thus, seven trials were retained for slope descent.
Along with the 3D motion capture data, the ground reaction forces (GRF) were also measured using two force plates (9287A, Kistler Group, Switzerland) at a frequency of 1000 Hz. 
For each trial, data from one gait cycle, defined by two consecutive ipsilateral heel contacts, was considered for further analysis. 

\subsection{Feature Extraction}

The biomechanical data acquired from the motion capture experiments were used to generate the joint angles and moments relevant for our study.
An open-source biomechanical modeling, simulation and analysis software, Opensim \cite{delp2007opensim} was used for this analysis. A generic model \cite{delp1990interactive} available in the Opensim repository was selected.
The scaling tool in Opensim was used to scale the generic model to fit the subject. 
Further, the scaled model and the marker trajectories were fed into the inverse kinematics (IK) tool of Opensim which solved an optimization problem to find the joint angles from the marker trajectories. The corresponding angular velocity and angular acceleration of hip and knee were obtained using successive numerical differentiation. The results of the IK, along with the recorded GRF were fed as input to the inverse dynamics (ID) tool of Opensim to obtain the corresponding joint moments. The IK and ID obtained joint angles and moments were considered as the ground truth for our study. 

\subsection{Shared Neural Network Model}

A three-layered fully connected feed-forward artificial neural network with 100 neurons each
was used as the model for continuously estimating the ankle joint angles and moments from the hip and knee joint kinematics. The model learns a shared representation of the different tasks (walking, stairs climbing, slope walking) without explicitly classifying them. 

The network was trained using a back-propagation algorithm for 30 epochs. A stochastic gradient descent (SGD) \cite{kiefer1952stochastic} with a momentum of $0.9$ was used as the optimizer.
An $l_2$ regularization penalty of $0.01$ and a learning rate of $0.0001$ were used for the weight updates with the mean-squared error as the loss criterion.
\subsection{Input and Output variables}

Time series sequences of the angles, angular velocities, and accelerations of hip and knee from the different locomotion modes, (level ground walking at self selected comfortable speed, slope ascent, slope descent, stair ascent and descent) were fed as input, $\textbf{x}$ to the model. The corresponding ankle angle and ankle moment were the output, $\textbf{y}$, of the model. 

\begin{equation}
\begin{aligned}
    \textbf{x} = 
    \begin{bmatrix}
        \theta_{hip}\\
        \dot{\theta}_{hip}\\
        \ddot{\theta}_{hip}\\
        \theta_{knee}\\
        \dot{\theta}_{knee}\\ \ddot{\theta}_{knee}
    \end{bmatrix} &, &
    \textbf{y} = 
    \begin{bmatrix}
        \theta_{ankle}\\
        \tau_{ankle}
    \end{bmatrix}
\end{aligned}
\label{eqn:inputs}
\end{equation}

Each input feature was low pass filtered using a 
Butterworth filter of 6Hz cut-off frequency  because our preliminary Fourier analyses revealed that at least 95\% of the signal information from the different kinematic signals used as input features are contained within 6Hz.
The input features were then normalized within the range 0 to 1. The final dataset to be used for model training and validation was of size 5000x6.

\subsection{Training and Validation of the Model}

The shared model was trained to predict the ankle angles and moments from the angles, angular velocities and accelerations of hip and knee at each time point. The relation learnt by the model is given by:
\begin{equation}
    y = f(x; w,b)
\end{equation}

where $x$, and $y$ are the inputs and outputs of the model as stated in eqn.(\ref{eqn:inputs}), and $w$, and $b$ are the parameters of the model.

The shared model was trained and validated using a leave-one-out (LOO) cross-validation strategy, in which during each iteration, data from a single trial of any of the locomotion mode was held out and the model was trained using data from all the remaining trials. The LOO was used to leverage the amount of data efficiently and to estimate the variance of the performance on the test set. The trained model was then used to predict the output for the held-out trial. This process was repeated until each of the trial was held-out once.
Two commonly used performance metrics, the coefficient of determination or the $R^2$ score and rms error (\textit{RMSE}) were used to evaluate the performance of the shared model.

\section{Results}

\subsection{Out-of-sample performance}
\begin{figure*}[!htbp]
    \centering
    \includegraphics[width=1\linewidth]{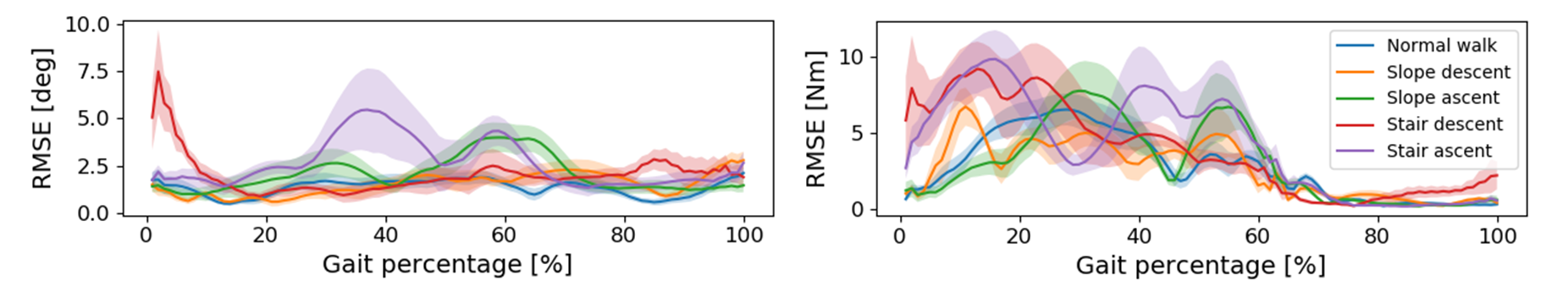}
    \caption{Mean absolute error of $\theta_{ankle}$ and $\tau_{ankle}$ prediction as a function of gait phase for different locomotion modes, colour-coded by gait types. Shaded area represents the inter-trial standard error. 0-60\% represents stance phase and 60-100\% represents swing phase of the gait. }
    \label{fig:gait_specific_error}
\end{figure*}
Fig. \ref{fig:gait_specific_error} shows the mean absolute error of predicted $\theta_{ankle}$ and $\tau_{ankle}$ as a function of the gait phase for different locomotion modes colour-coded by gait types. It was observed that for $\theta_{ankle}$ predictions, high mean absolute errors occurred during initial stance for stair descent, mid-stance and stance-to-swing transition for stair ascent, and stance-to-swing transition for slope ascent. The mean absolute error of $\theta_{ankle}$ for normal walk and slope descent remained below 3\degree\ during the complete gait cycle. For $\tau_{ankle}$ predictions, errors occurred mostly in the stance phase for all the locomotion modes, where the desired $\tau_{ankle}$ is non-zero.  Mean absolute errors for normal walk and slope descent were lower compared to other locomotion modes. Errors in stair descent occurred mainly during start of the stance phase. Errors in slope ascent occurred during mid- and terminal- stance whereas in stair ascent during initial and terminal stance. 

In the appendix we also show, in Figure \ref{fig:r2_rmse}, the $R^2$ and the RMSE plots for quantifying the performance of the shared model for the different gait types.  
\subsection{Comparison with Baselines}

We compared the performance of our shared neural network-based model against two baselines: a linear model as well as a non-linear model, (support vector regression with \textit{rbf} kernel). The results are as summarized in Table \ref{table:baseline_comp}. The comparisons were quantified using the evaluation metrics, $R^2$ score and the rms error (\textit{RMSE}) which showed that our model performs better across the different gait types than the baseline reported results.

\begin{table*}[!htbp]
{ \centering
\begin{tabular}{ccccccccccccc}
    & \multicolumn{6}{c}{Linear Model}                                                       & \multicolumn{6}{c}{Support Vector Regression}                                          \\
    & \multicolumn{2}{c}{StairDescent} & \multicolumn{2}{c}{StairAscent} & \multicolumn{2}{c}{Normal Walk} & \multicolumn{2}{c}{StairDescent} & \multicolumn{2}{c}{StairAscent} & \multicolumn{2}{c}{Normal Walk} \\
    & $\theta$       & $\tau$       & $\theta$      & $\tau$      & $\theta$     & $\tau$    & $\theta$       & $\tau$       & $\theta$      & $\tau$      & $\theta$     & $\tau$    \\
$R^2$  & 0.33           & 0.36         & 0.35          & 0.50        & 0.42         & 0.74      & 0.95           & 0.88         & 0.86          & 0.93        & 0.94         & 0.96      \\
RMSE & 15.43          & 17.27        & 9.19          & 17.41       & 7.75         & 14.91     & 3.88           & 7.06         & 4.09          & 6.46        & 2.44         & 5.07     
\end{tabular}
}
\vspace{-0.3 cm}
\caption{Experiments for baseline comparison. The results are shown on the test set for 3 modes.}
\label{table:baseline_comp}
\end{table*}

\section{Discussion and Conclusion}

In this study we proposed a shared neural network model to predict the ankle joint kinematics and kinetics across varied locomotion tasks without explicitly classifying them. The results of our study suggest that by using the hip and knee joint kinematics as inputs, it is possible to estimate the target variables with high accuracy across the different tasks. 
The shared model could predict the ankle angle and moment on the held-out trial with mean $R^2$ greater than 0.89. For $\theta_{ankle}$ prediction, highest mean $R^2$ was obtained for stair descent and for $\tau_{ankle}$ prediction, highest mean $R^2$ was obtained for level ground walking data. We compared the proposed model with prior work on separate task models. We found that our model could generalize better to the array of different tasks. Furthermore, using separate task models require a two-stage model where first the mode is detected (walking, stair climbing etc.) and then, the angles and moments are estimated. This approach is more complex and prone to error accumulation. When errors or latency occur in the mode classification the performance of the two-stage approach would deteriorate.

In a real world scenario, our algorithm could be used as a high-level controller to generate set-point commands for controlling a powered prosthetic/orthotic ankle across different locomotion modes. The inputs could be acquired from one or multiple combinations of miniaturized wearable inertial sensors (IMUs) \cite{o2007inertial}, wearable goniometers, stretch sensors \cite{tognetti2015wearable, mattmann2008sensor} as opposed to motion capture equipment which is limited to laboratory conditions. 
This approach could be a first step towards the development of context-aware high level control strategies which can seamlessly and continuously adapt to different locomotion modes without identifying gait phases or switching between mid-level controllers. 




\typeout{} 
\bibliography{jmlr-sample}
\newpage
\newpage
\onecolumn
\appendix
\counterwithin{figure}{section}
\counterwithin{table}{section}


\section{}\label{apd:first}

\begin{figure*}[!htbp]
    \centering
    \includegraphics[width=1\linewidth]{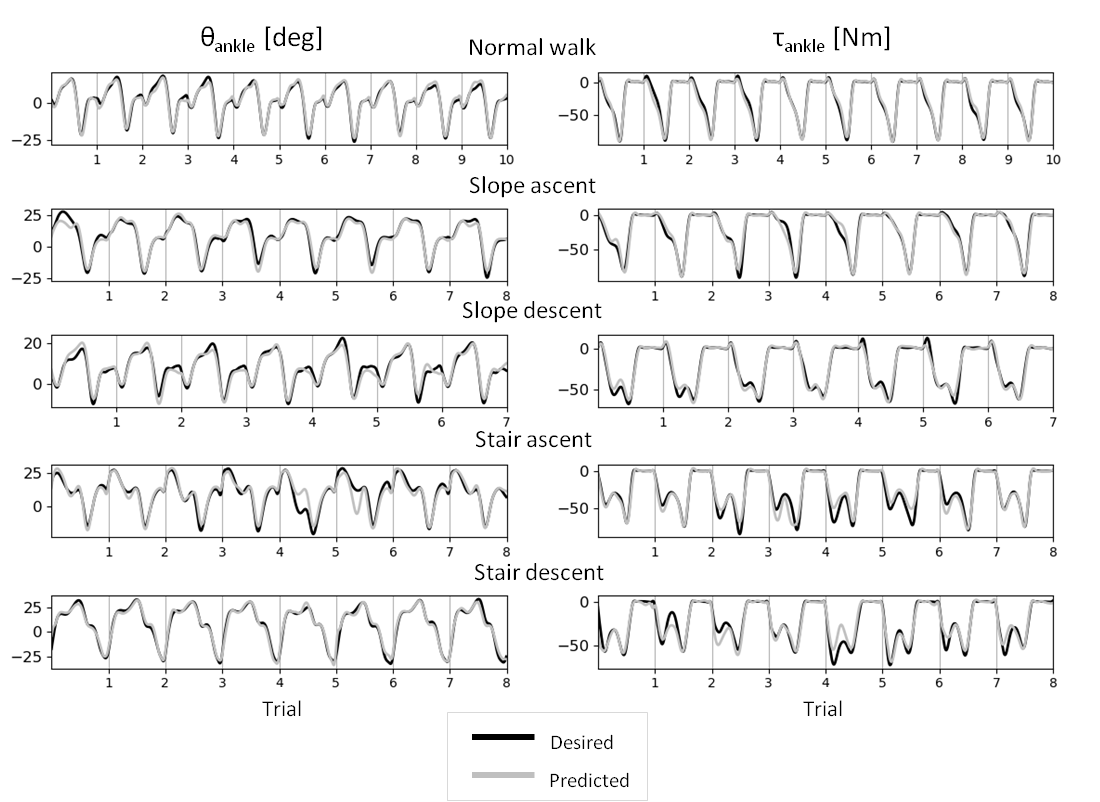}
    \caption{Expected (black) and predicted (grey) values of ankle angle $\theta_{ankle}$ and ankle torque $\tau_{ankle}$ for different trials of normal walking, slope ascent, slope descent, stair ascent, and stair descent. Each trial show a distinct gait cycle. }
    \label{fig:prediction}
\end{figure*}

\begin{figure*}[!htbp]
    \centering
    \includegraphics[width=1\linewidth]{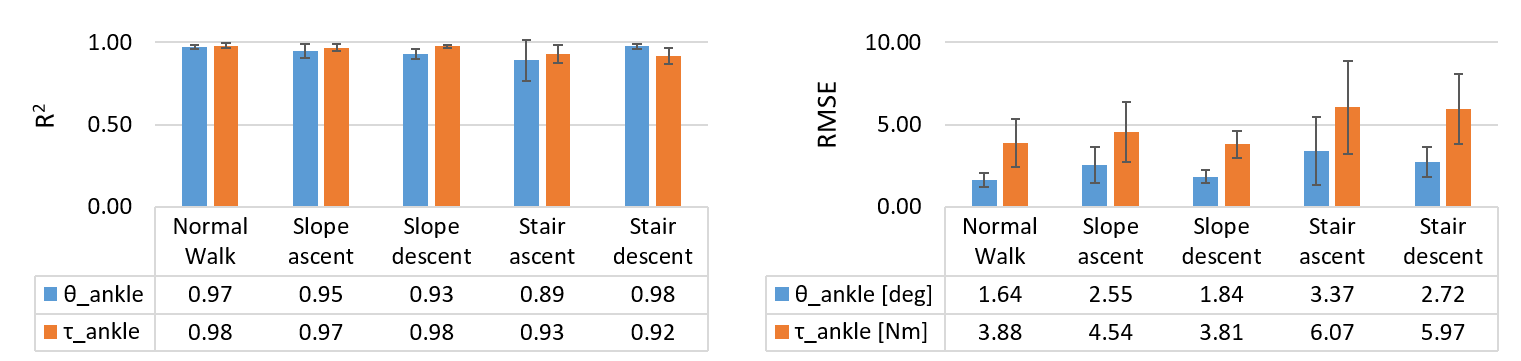}
    \caption{Mean $R^2$ scores and \textit{RMSE} of estimation of ankle angle and ankle moment using the proposed model on datasets recorded from different locomotion modes. Error bars show inter-trial standard deviation.}
    \label{fig:r2_rmse}
\end{figure*}

Fig. \ref{fig:r2_rmse} shows the mean $R^2$ and \textit{RMSE} of the shared model in estimating the ankle angle and ankle moment during different locomotion modes. The proposed  model achieved mean $R^2$ scores between 0.89 (stair ascent) and 0.98 (stair descent) for $\theta_{ankle}$ and between 0.92 (stair descent) and 0.98 (normal walk and slope descent) for $\tau_{ankle}$. An \textit{RMSE} between 1.64\degree\ (normal walk) and 3.4\degree\ (stair ascent) was obtained for $\theta_{ankle}$ and between 3.81 Nm (slope descent) and 6.1 Nm (stair ascent) for $\tau_{ankle}$ estimation. From the \textit{RMSE} values, it was observed that the highest overall prediction performance was obtained for normal walk and slope descent followed by slope ascent and finally stair descent and ascent.
Fig. \ref{fig:prediction} shows the desired and predicted values of ankle angle and ankle torque for different locomotion modes. 
It was observed that the predicted trajectories of $\theta_{ankle}$ and $\tau_{ankle}$ closely followed the desired trajectories for all the normal walking trials. The \textit{RMSE} for individual walking trials ranged between 1.12\degree and 2.31\degree for $\theta_{ankle}$ and between 1.77Nm and 6.91Nm for $\tau_{ankle}$. Similarly, the $\theta_{ankle}$ and $\tau_{ankle}$ predictions for slope ascent trials closely matched the desired trajectory, with small deviations in some trials. The \textit{RMSE} for individual trials ranged between 1.17\degree and 4.55\degree for $\theta_{ankle}$ and between 1.44Nm and 7.11Nm for $\tau_{ankle}$. The $\theta_{ankle}$ predictions for slope descent showed small deviations during the terminal swing phase for some trials, but the $\tau_{ankle}$ predictions were close to expected across trials. The \textit{RMSE} values ranged between 1.24\degree and 2.32\degree for $\theta_{ankle}$ and between 2.88Nm and 5.38Nm for $\tau_{ankle}$ across slope descent trials. The $\theta_{ankle}$ predictions for stair ascent and stair descent closely followed the desired trajectory except for two stair ascent trials. Correspondingly, the \textit{RMSE} values of $\theta_{ankle}$ ranged between 1.71\degree and 7.25\degree for stair ascent and between 1.73\degree and 4.45\degree for stair descent. The $\tau_{ankle}$ predictions for stair ascent and descent deviated from the desired trajectory during a few trials. The \textit{RMSE} values ranged between 1.63Nm and 9.98Nm for stair ascent and between 2.39Nm and 9.95Nm for stair descent. 
\end{document}